\documentclass[letterpaper, 10 pt, conference]{ieeeconf}  % Comment this line out if you need a4paper

\IEEEoverridecommandlockouts                              % This command is only needed if 
\usepackage{comment}
\usepackage{caption}
\usepackage{subcaption}
\usepackage{amsmath,amsfonts,amssymb}
\usepackage{multirow}
\usepackage{array}
\usepackage{gensymb}
\usepackage{float}
\usepackage{textcomp}
\usepackage{graphicx}
\usepackage{algorithm}
\usepackage{algpseudocode}
\usepackage{epstopdf}

\usepackage{manfnt}

\usepackage{cite}
\usepackage[colorlinks=true, allcolors=blue]{hyperref}
\usepackage{hyperref}
\usepackage{xfrac}
\usepackage{mathtools}

\title{\LARGE \bf
User-Driven Learning from Demonstration: A Trajectory and Impedance Learning Method}

\author{Zi-Qi Yang  
and Mehrdad R. Kermani
\thanks{Authors are with the Department of Electrical and Computer Engineering, Western University, London, ON N6A 5B9, Canada.
        {\tt\small Email: zyang524@uwo.ca, mkermani@eng.uwo.ca}}%
}

\begin{document}

\maketitle
\thispagestyle{plain}
\pagestyle{plain}

\begin{abstract}
This paper presents a method for user-driven robot Learning from Demonstration (LfD) that reduces user effort while ensuring compliant and precise reproduction. The method eliminates repeated teaching for the same task and enables real-time learning from a single demonstration. Demonstrated motions are reproduced with high precision, while impedance variations are learned in real time to provide both compliance and robustness against perturbations. This mitigates potential safety issues in Human-Robot Interaction (HRI) that arise from conventional time-indexed trajectories lacking compliance. The proposed approach integrates a three-dimensional (3D) Fast Diffeomorphic Matching (FDM) algorithm with a Dynamical System (DS)-based motion generator to achieve real-time single-shot demonstration learning and reproduction. An Extended Kalman Filter (EKF) framework compensates for reproduction errors and recovers from external interactions. Furthermore, an impedance parameterization function is incorporated to learn impedance variations from demonstrations and maintain surface contact for specific applications. 
The proposed approach is validated through comprehensive experiments on a 7 Degree-of-Freedom (DOF) KUKA LWR IV+ robot. 
% \footnote{Video of the experiments can be found in the supplementary material.}

\end{abstract}

\section{Introduction}
Industrial robots are increasingly
deployed to reduce human labour in tedious, repetitive tasks. Despite advances in machine learning and AI, Human-Robot Collaboration (HRC) has remained a critical   necessity since robots cannot match human capabilities in terms of variability and task diversity. These tasks usually require human involvement to provide supervision, demonstrations, and modifications to robot motion \cite{mukherjee2022survey}, during object handling, collaborative manufacturing and assembling, and others. Learning from Demonstration (LfD) is one of the promising approaches \cite{argall2009survey, ravichandar2020recent} to cope with these task requirements that increase efficiency and flexibility of HRC tasks.

Specifically, among many approaches for demonstrations, kinesthetic teaching provides the most straightforward way. The recorded demonstrations are first preprocessed to rule out undesired demonstrations by methods including filtering demonstrated data \cite{coates2008learning}, identifying suboptimal data \cite{choi2016robust}, and quantifying demonstration
quality \cite{sakr2022quantifying}. More recent research also shows the importance of maintaining consistent interaction feeling to the demonstration quality \cite{yang2026user}. 
% the existing LfD approaches often utilize sensors to provide guidance such as cameras to help with the precision of human demonstration, as well as .  While these methods excel within their respective domains,
Subsequently, for feature extraction, the satisfied demonstrations are learnt mainly by two broad categories of methods (i) deterministic approaches that try to approximate the demonstration with an explicit function, e.g., a radial basis function network, and (ii) probabilistic approaches that learn a probability distribution from the demonstrations. The first category of approaches usually requires offline training and high computational complexity, and thus is ruled out of consideration for most instant LfD tasks. The second category of approaches can provide satisfactory reproductions, most of the existing probabilistic methods are based on statistical data from multiple demonstrations to reproduce the ideal trajectories, and the replicated trajectories are with explicit time dependency. 

In the literature, three models are mainly used to extract and generalize information from multiple demonstrations,
namely the probabilistic-based hidden Markov model (HMM), Gaussian mixture model (GMM), and dynamic systems (DS)-based Dynamic movement primitives (DMPs). HMMs are a popular probabilistic methodology to encode the demonstrations, which captures both spatial and temporal variability by viewing a demonstrated trajectory as a sequence of latent motion states with probabilistic transitions \cite{lafleche2018robot}.
%They are suited to segmenting complex skills, recognizing phases online and reproducing motions that respect the demonstrated timing structure.
GMMs provide a probabilistic encoding of the observed data distribution \cite{feng2025improved}. When combined with Gaussian mixture regression (GMR), they yield generalized reference trajectories and an explicit covariance that can adapt compliance during task execution \cite{wang2021optimised}.
Complementing these mixture-based approaches, Gaussian Process Regression (GPR) offers a non-parametric, kernel-based alternative that interpolates smoothly within the neighbourhood of the demonstrations while simultaneously returning a predictive variance \cite{franzese2021ilosa}.
When reproducing, these approaches do not allow adaptation to new trajectories, as it is identified as perturbations that deviate from the task reference, unless another demonstration procedure is provided.

DMPs model each movement as a stable, nonlinear DS whose parameters are learned from demonstration. 
DMPs are time-invariant, guarantee convergence to a goal, and can be reshaped on the fly, making them suitable to be applied to robot movements \cite{chen2017robot}.  Although these methods can be adapted and combined with other approaches to help improve the efficiency and flexibility in learning the trajectories, the multiple demonstrations and time invariant characteristics remain largely unsolved. Recently, the DS also emerged as a ﬂexible means of representing robot motions independent of time \cite{yang2025nullspace}.

Additionally, the force and impedance information from demonstrations provides crucial information especially in contact tasks and constrained environment, which suggest this information must be embedded directly into the learning process. Existing methods for integrating the force/impedance information are usually rely on sensor feedback. These approaches usually provide time-indexed measurement and are recorded independently of the path reproduction. Recently, force/impedance information has been incorporated into path reproduction models to enable compliant motion reproduction \cite{ge2024learning}. However, such approaches typically rely on simplified environment contact models and cannot be directly learned from human demonstrations.

\subsection{Contribution}
To overcome these challenges, 
and extend our prior work \cite{yang2026user,yang2025nullspace,yang2023computationally,kermani2023antagonistic,yang2026unified}, 
we aim to develop a single-shot LfD framework that simultaneously acquires positional and impedance profiles directly from the robot’s proprioceptive sensors, then reproduces the demonstrated paths with high fidelity while maintaining contact across varying surface geometries. Our proposed approach builds on a linear DS that is transformed into a nonlinear Cartesian space translational motion generator via a 3-dimensional (3D) Fast Diffeomorphic Matching (FDM) algorithm. 
Reproduction accuracy is substantially improved by combining an Extended Kalman Filter (EKF) with the proposed FDM-based motion generator.
Inspired by the impedance parameterization method in \cite{kronander2015passive}, we encode direction-dependent damping online along the learned paths. 
Finally, using the velocity direction estimate, we generate a velocity direction along the surface normal of the taught trajectory, enabling compliant yet reliable surface contact. Summarizing, the main contributions of this paper are: 
\begin{itemize}
    \item A 3D FDM-based algorithm integrated with a DS framework that enables learning from single-shot demonstrations and reproduces them in real time.
	\item An EKF framework-based velocity correction module that achieves centimeter-level reproduction accuracy, enabling rapid recovery from external perturbations.
    \item A real-time impedance learning method that selectively regulates interaction dynamics to resist potential deviations induced by external contacts along the path, while maintaining surface contact when required.
\end{itemize}

The remainder of this paper is organized as follows. Section~\ref{method} presents the proposed methodology, Section~\ref{exp} describes the experimental evaluation, and Section~\ref{Conclusion} concludes the paper.

\section{Methodology}
\label{method}
In this work, we consider an $n$--DOF redundant serial torque-controlled robot manipulator. The dynamic model of the robot in joint space can be expressed as, 
\begin{equation}
\label{dynamicmodel}
M(q)\ddot q + C(q,\dot q)\dot q + G(q) = \tau_{c}  + \tau_{f} + \tau_{ext},
\end{equation}
where ${q \in \mathbb{R}^{n} }$ denotes the joint configuration, ${M(q) \in \mathbb{R}^{n\times n} }$ is the inertia matrix,  ${C(q,\dot q) \in \mathbb{R}^{n \times n} }$ represents the Coriolis and centrifugal matrix, ${G(q) \in \mathbb{R}^{n} }$ denotes the gravitational torques, whose effect can be accurately compensated in most
industrial robot’s internal controllers. $\tau_{c}\in \mathbb{R}^{n}$ denote the control torques produced by our proposed method. $\tau_{f}\in \mathbb{R}^{n}$ denotes the joint frictional torques, and
% ${\tau_{c} = \tau_{ns}+ \tau_{cart} + \tau_{f} \in \mathbb{R}^{n} }$ denotes the joint control torque to be designed,
${\tau_{ext}\in \mathbb{R}^{n }}$ denotes the unknown external interaction torque exerted on the end-effector (EE) that reflected on robot joints.
% In the following, the dependency on joint configuration $(q, \dot q)$ is omitted for brevity. 

\begin{figure} [!ht]
   \begin{center}
\includegraphics[width=0.5\textwidth]
{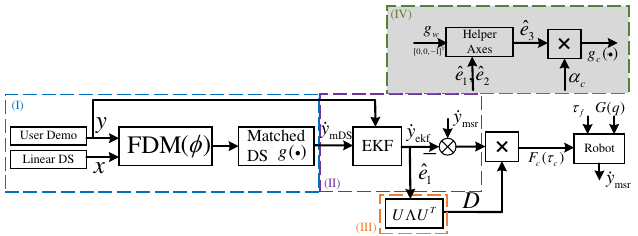}
   \end{center}
   \caption{Overview of the proposed framework architecture. (I) Demonstration path matching
   and velocity generation; (II) velocity correction and perturbation recovery; (III) impedance learning; and (IV) an optional module for maintaining surface contact.}
   \label{Control_Dia}
\end{figure}

The architecture of the proposed framework is presented in Fig. \ref{Control_Dia}.  (i) A linear DS is matched to the demonstrated trajectories using the 3D FDM algorithm, yielding a base velocity field for motion tracking; (ii) the base velocity is subsequently corrected via an EKF to refine trajectory reproduction, and enables perturbation recovery; (iii) impedance parameters are learned online during the reproduction process; (iv) an optional module is incorporated to maintain  surface contact when required.  This work aims to lighten the burden of human workers in repetitive and tedious kinesthetic teaching tasks by learning positional and the corresponding impedance profiles from single-shot demonstrations. The proposed approach enables robots to accurately reproduce demonstrated motions with adjustable compliance, remaining responsive to external interactions and, when required, maintaining contact with different target surfaces.
In the following, we explain each part of the framework in detail.

\subsection{Motion Generation}
The motion generation module is composed of three interconnected submodules, namely the FDM for matching the demonstrated path, the matched DS for generating initial velocity, and the EKF for correcting the velocity estimation errors in real time. 
\subsubsection{Fast Diffeomorphic Matching}
The FDM method, introduced in \cite{perrin2016fast}, aims to compute a diffeomorphism $\Phi$ that maps each point $x_i \in X \subset \mathbb{R}^2$ onto $y_i \in Y\subset \mathbb{R}^2$, where $X = (x_i)_{i\in\{1,...,N\}} $ and $Y = (y_i)_{i\in\{1,...,N\}} $ are two sequences of distinct points. 
In this work, we extend the application of this method to 3D, where $X \subset\mathbb{R}^3$ and $Y \subset\mathbb{R}^3$. Each point in $X$ is transformed
using the locally weighted translation, 
\begin{align}\label{FDM}
  \hat y_i = \phi_{\rho_j, c_j,v_j}(x_i) = x_i + k_{\rho_j}v_j,
\end{align}
where $\hat y_i$ denotes the translated point in $\hat Y\in \mathbb{R}^3$, $c_j\;\text{and}\; v_j \in \mathbb{R}^3$ denotes the center and direction of the $j^{th}$ translation. A smooth diffeomorphism is the composition
of all single locally weighted translations. In which, $k_{\rho_j} = e^{-\rho_j^2||x_i-c_j||^2},$
is a Gaussian Radial Basis Function (RBF) kernel, where $\rho_j$ represents the coefficient to be optimized to minimize the distance between $\phi_{\rho_j, c_j,v_j}(X)$ and $Y$, \begin{align}\label{rho}
   & \rho_j := \min \,\, \frac{1}{N}\sum_j ||\phi_{\rho_j, c_j,v_j}({X}) - Y||^2,
\end{align}
where $N$ denotes the number of points in $X$ and $Y$, and $\rho_{\text{max}}=\frac{e^{ (\sfrac{1}{4})}} {\sqrt{2}||v||}$ as the maximum value for better mapping.

In the following, we briefly describe the FDM algorithm (Algorithm 1). Firstly, we select the point $x_m$ in $X$ that is the furthest from its corresponding target $y_m$ in $Y$. Here, the point set $X$ can be a straight line connecting the start and the end of the demonstrated path $Y$.  Then, we optimize the $\rho_j$ to minimize the distance between 
$\phi_{\rho_j, c_j,v_j}({X})$ and $Y$ as in \eqref{rho}. In which, 
the direction of the $j^{th}$ translation $\phi_{\rho_j, c_j,v_j}$ is updated to $v_j = \beta (q_j - c_j)$, with $0<\mu<1$ representing the safety margin, $0<\beta\leq1$ denoting the learning rate, and $K=120\text{–}200$ is the total number of locally weighted translations selected by the user, where the loop iterates over each translation. $\lambda$ term denotes a regularization term. Finally, we update the estimated demonstration path $\hat Y$  by compositing  the $K$ locally weighted translations together as $\hat{Y}=\Phi( X^{(K)})=\bigl(\phi_{\rho_1,c_1,v_1}\circ\phi_{\rho_{2},c_{2},v_{2}}\circ\cdots\circ\phi_{\rho_K,c_K,v_K}\bigr)( X)$.

% ─── Algorithm I ─────────────────────────────────────────────────────
\begin{algorithm}[t]
\caption{3D FDM of $\hat{Y}=\Phi(X)$}
\label{alg:I}
\begin{algorithmic}[1]
\Require $ X=(x_i)_{i\in\{1,\dots, N\}}$,\; $ Y=(y_i)_{i\in\{1,\dots, N\}}$\\
$K\in\mathbb N_{>0},\; 0<\mu<1,\; 0<\beta\le 1$
\State ${\hat{Y}}  = (\hat{y}_i)_{i\in\{1,\dots, N\}} = 0$ \Comment{init}
\For{$j = 1$ \textbf{to} $K$}
    \State $m = \displaystyle\arg\max_{i\in\{1,\dots,N\}}
                 \bigl\lVert {x}_i - y_i \bigr\rVert$
                 \Comment{find furthest point$\#$}
    \State $c_j = {x}_{m}$ \Comment{furthest point}
    \State $q_j    = y_{m}$ \Comment{demo point}
    \State $v_j = \beta\,(q_j - c_j)$ \Comment{direction of $\phi$}
    \State $\displaystyle
           \rho_j = \displaystyle\underset{\rho\in[0,\mu\rho_{\max}(v_j)]}{\arg\min}\! \frac{\sum_i^N||\phi_{\rho,c_j,v_j}({{X}})-\,
                                 Y||^2}{N}+\lambda \rho_{max}^2$
     \State${\hat{Y}} = \phi_{\rho_j,c_j,v_j}({{X}})$      
     \Comment{composition of all}
     \EndFor \\
    \Return $\bigl\{q_j,\rho_j,c_j, v_j\bigl\}_{j\in\{1,\dots, K\}}$
\end{algorithmic}
\end{algorithm}
% ─────────────────────────────────────────────────────────────────────

\subsubsection{Motion Reproduction}
\label{Motion Reproduction}
To reproduce the demonstrated paths matched by the FDM algorithm, this subsection first illustrates the FDM algorithm's ability to learn globally asymptotically stable (GAS) nonlinear DS that functions as a motion generator. The literature has shown that DS-based motion representations possess rich real-world applications and can be rigorously proven to be GAS in \cite{khansari2011learning}, which is one of the many benefits of our proposed motion reproduction approach. The authors in \cite{perrin2016fast} utilize the derived diffeomorphism $\Phi$ to map a linear DS formulated as $\dot{x} = -x$ onto the demonstrated paths in 2D space with single motion pattern. The initial and mapped DS are formulated as,
\begin{align}
   &\dot{x}= -\gamma(x)x,\label{LinearDS}\\
   &\dot{y}= -\gamma(\Phi^{-1}(y))J_{\Phi}(\Phi^{-1}(y))\Phi^{-1}(y),\label{LinearDS2}
\end{align}
where  the velocity tuner is defined as $\gamma(\cdot) = \frac{||y_1||}{\delta tN||x||}\in \mathbb{R}^3 \rightarrow \mathbb{R}^+ \;\text{for}\; ||x||\geq \frac{||y_1||}{N}$ and $\gamma(\cdot) = \frac{||y_1||}{N}$ otherwise, $y_1$ denotes the first element. In \eqref{LinearDS}, \eqref{LinearDS2}, the subscripts of individual points $x_i\in X$ and $y_i \in Y$ are omitted for brevity, and $\dot x$, $\dot y\in \mathbb{R}^3$ represent the produced velocity by the linear and FDM-transformed DS, respectively.

Although FDM (Algorithm 1) can successfully match demonstrated paths in both 2D and 3D, the motion generated by the transformed DS in \eqref{LinearDS2} deviate substantially from the original demonstrations.  Its generalization capability deteriorates in higher-dimensional (3D) settings and for more complex motion cases. To solve this problem, we continue in 3D space and propose a modified DS (mDS) to significantly improve the original DS's ability to reproduce demonstrations. The mDS-based motion generator is,
\begin{align}\label{DS}
   \dot y_{\text{mDS}}&= -\zeta\,J_{\Phi}\,\hat x,
\end{align}
where $\dot y_{\text{mDS}}\in \mathbb{R}^{3}$ represents the velocity generated by the mDS in translational space, $\zeta= \zeta\, + \, \eta\,\Delta \zeta \in \mathbb{R}^{3\times 3}$ is a velocity modulation term. In which, $\Delta\boldsymbol{\zeta} =  \frac{
(\cdot)^T {e}_v
}{\left\| (\cdot)^T {e}_v \right\| + \epsilon}$, with ($\cdot$) denotes the partial derivative $\frac{\partial {\dot y_{\text{mDS}}}}{\partial \boldsymbol{\zeta}}$.   $\hat x \in \mathbb{R}^3$ denotes the individual point estimated from $\hat{X}=\Phi^{-1}(Y)$.
$e_v \in \mathbb{R}^3$ denotes the velocity error between the original demonstration and mDS generated velocity. $\eta$ and $ \epsilon \in \mathbb{R}^+$ denote the adaptation rate and threshold, respectively. 
In practice, $\Delta\zeta$ is filtered by a moving average filter and applied with clamping.

In the following, we derive the inverse of the diffeomorphism (Algorithm 2) for $\hat x$. It is denoted as  $\hat X=\Phi^{-1}(Y)$, and is obtained by reversing the displacement in \eqref{FDM} as, 
\begin{align}\label{X = Y-k}
\Phi^{-1}(Y) = Y+{k}_j v^-_j, 
\end{align}
where $v^-_j = -\beta (q_j - c_j)$ denotes the negative direction of $v_j$. The RBF weight at iteration $j$ is
\begin{align}
\label{eq:rbf}
k_j = e^{\left(-\rho_j^2 \|\hat{X} - {c}_j\|^2\right)}.
\end{align}
Letting $o_j:=Y-c_j$, we can rewrite \eqref{eq:rbf} as,
\begin{align}\label{lambda}
r_j(k) := k_j = e^{\left(-\rho_j^2 \|{o_j} + k_j v^-_j\|^2\right)}, \quad k_j\in[0,1].
\end{align}
Since  \eqref{lambda} is a self-referential equation that cannot be solved algebraically, we therefore rewrite it as a 1D root finding problem $F_j(k) := r_j(k) - k = 0$ and apply Newton–Raphson method, which exhibits local quadratic convergence under standard regularity conditions.
Differentiate $F_j(k)$ as,
\begin{align}\label{F_deriv_1}
F_j'(k) = \frac{d}{d\,k}[r_j(k)]-1.
\end{align}
To solve this, let \(* \coloneq -\rho_j^2 \|{o_j} + k_j {v}^-_j\|^2\), and write the first term  as,
\begin{align}
\frac{d}{d\,k}\left[e^{(*)}\right] = e^{(*)} \frac{d *}{d\,k},
\end{align}
where $e^{(*)}$ denotes the exponential function with argument $*$. Then calculate the derivative of \(*\) as,
\begin{align}
\frac{d *}{d\,k} = -2\rho_j^2 ({o_j} + k_j {v}^-_j)^T {v}^-_j,
\end{align}
since \(\frac{d}{d\,k}\|{o_j} + \,k_j {v}^-_j\|^2 = 2({o_j} + \,k_j {v}^-_j)^T {v}^-_j\). Thus,
\begin{align}
F_j'(k) = e^{(*)} \cdot \left[-2\rho_j^2 ({o_j} + k_j {v}^-_j)^T {v}^-_j\right] - 1.
\end{align}
Rewriting in terms of \(r_j(k) = e^{(*)}\),
\begin{align}
F_j'(k) = -2\rho_j^2 r_j(k) \left[ ({o_j} + k_j {v}^-_j)^T {v}^-_j \right] - 1. 
\end{align}
Finally, the update of $k_j$ can be written as,
\begin{align}
k_{{j+1}} = k_j - \frac{F_j(k)}{F_j'(k)}.
\end{align}

The detailed procedure is summarized in Algorithm 2, where the symbols defined in Algorithm 1 are inherited and parentheses are omitted for simplicity. In practice, the stopping threshold $\epsilon$ is set to a small positive value (e.g., $10^{-6}$), and a maximum number of iterations (e.g., 20–50) is imposed to avoid potential non-termination.
\begin{algorithm}[!ht]
\caption{Newton method of $\hat{X}=\Phi^{-1}(Y)$}
\label{alg:phiInv}
\begin{algorithmic}[1]
\Require
  $ Y=(y_i)_{i\in\{1,\dots, N\}}$, $\bigl\{ c_j, q_j, \rho_j\bigr\}_{j=1}^K$\\
  $K\in\mathbb N_{>0}, 0<\beta\le 1, \epsilon = 10^{-6}$
\For{$j = K \ \textbf{to}\ 1$}               \Comment{undo $\Phi_K\circ\dots\circ\Phi_1$}
    \State $\displaystyle  v^-_j = - \beta({ q_j- c_j}){}$  \Comment{reversed direction}
    \State \( e_j   =  Y- c_j\)
        % \State $k_j= {0}_N\in \mathbb{R}^{1\times N}$              \Comment{initial root}
        \Repeat                                   \Comment{Newton method}
            \State $\displaystyle 
                 r_j  =
                 e^{\!\bigl(-\rho_j^{2}\odot\,
                    \lVert o_j+k_j\odot v^-_j\rVert^{2}\bigr)}$\Comment{kernel}
            \State $g_j = (2[ o_j +k_j\odot v^-_j)\;{\!\odot}\; v^-_j]^T {1}_N$ \Comment{${1}_N\in\mathbb{R}^{N \times 1}$}
            \State $F_j = r_j-k_j,\qquad
                   F' = -1-\rho_j^{2} \odot g_j\odot r_j$
            \State $\displaystyle
                   k_{j+1} = k_{j}-F_j\oslash F_j'$
        \Until{$|F_j|<\epsilon$}\Comment{threshold}
        \State $ \hat{ X} =  Y+k_j\odot v^-_j$ \Comment{output}
    \EndFor
\end{algorithmic}
\end{algorithm}

Given that many variables are already derived in the Algorithm 2, it is natural and numerically efficient to formulate Jacobian matrix of the diffeomorphism in  \eqref{DS} as a Sherman–Morrison formula, $J_{\phi_j}(\hat{X}) = I_3 + u_j\,w_j^T,$
where $I_3 \in\mathbb{R}^{3\times3}$ is an identity matrix, 
and  $u_j:=-2\rho^2_j k_j v^-_j$, $w_j := \hat{X} - c_j$. {The Jacobian is written as,}
\begin{align}
  J_{\phi_j}=
I_{3}+{2\rho^2_j\,k_j}
       % {1-2\rho^2_j\,\lambda\, (v^-_j)^{\!T}(\hat{x}- p_j)}
       \, v^-_j\,(\hat{X}- c_j)^{\!T},
\end{align}
with its dependency $\hat{X}$ omitted. The full Jacobian $J_{\Phi}\in \mathbb{R}^{3\times 3}$ of the diffeomorphism
$\Phi$ is,
\begin{align}
J_{\Phi}=
\prod_{j=K}^{1}
  J_{\phi_j},
        \quad
j = K\rightarrow 1.
\end{align}

\subsubsection{Correction of Reproduced Motion under Perturbations}
Building upon the velocity generated by the mDS in \eqref{DS}, we further introduce a compensation mechanism to mitigate the amplified tracking errors that arise when reproducing complex motions in real-world environments.  Specifically, we leverage the Kalman gain scheduling property of the EKF framework to refine the velocity $\dot y_{\text{mDS}}$ produced by the mDS. This correction enables centimeter-level accuracy in motion reproduction, even under external perturbations. Our approach uses a low-dimensional EKF state, which is more compact than estimating full motion derivatives, yet sufficient to correct residual velocity errors.  We augment the mDS with an additional velocity bias term $b$, estimated online via the EKF. The corrected velocity is  given by,
\begin{align}\label{v_ekf}
    \dot y_{\text{ekf}} = \dot y_{\text{mDS}} + b.
\end{align}
 We define the EKF state as $s_i=[y_i,b_i]^T \in \mathbb{R}^{6}$, where $y_i\in \mathbb{R}^{3}$ is the $i^{th}$ Cartesian position in the  demonstration path point set $Y$, $b_i\in \mathbb{R}^{3}$ denotes the additive velocity bias term. 
 We then write the discrete-time process model as,
\begin{align}\label{processMod}
f_d(s_i)=
  \begin{bmatrix}
     y_i + \bigl(g(\cdot) + b_i\bigr)\,\Delta t + w_{y,i},\\[2pt]
   e^{-\lambda_b\Delta t}\,b_i + w_{b,i},
  \end{bmatrix},
\end{align}
where the mDS is denoted as $g(\cdot)$ and is defined as $g(\cdot):=\dot y_{\text{mDS}}$. $\Delta t$ is the sampling time, and
$\lambda_b > 0$ is the bias decay rate. The process noises are modelled as,
$
w_{y,i}\sim\mathcal N(0,Q_y\Delta t),
w_{b,i}\sim\mathcal N(0,Q_b\Delta t),
$ where $Q_y$, $Q_b\in \mathbb{R}^{3\times 3}$ are user-defined covariance matrices. 
Then, to update the covariance, 
we linearize \eqref{processMod} at the current estimate $\hat s_{i|i}=[\hat y_{i|i},\hat b_{i|i}]^T$. The state transition Jacobian is, 
\begin{align}
F_i := \left.\frac{\partial f_d}{\partial s}\right|_{s=\hat s_{i|i}}
= 
\begin{bmatrix}
I_3 + \,F_y(\hat y_{i|i})\Delta t & \,I_3\Delta t\\
0_3 & e^{-\lambda_b\Delta t}\,I_3
\end{bmatrix},
\end{align}
where $F_y(\hat y_{i|i}) := 
\left.\frac{\partial g(\cdot)}{\partial y}\right|_{y=\hat y_{i|i}}
\in\mathbb R^{3\times 3}$. 
In practice, $F_y$ can be approximated using  forward finite differences approach with a small time interval (e.g., $10^{-3}$). ${I}_3$ is the $3\times 3$ identity matrix, and ${0}_3$ is the $3\times 3$ zero matrix. Then,  the EKF recursion is given by the prediction step,
\begin{align}
\left\{
\begin{aligned}
\hat s_{i|i-1} &= f_d(\hat s_{i-1|i-1}),\\
P_{i|i-1} &= F_{i-1} P_{i-1|i-1} F_{i-1}^T + Q_{i-1},
\end{aligned}
\right.
\end{align}
where $P_{i|i-1} \in \mathbb{R}^{6\times 6}$ denotes the predicted covariance estimate, and
$Q = \text{blkdiag}(Q_y\;Q_b)\;\Delta t \in \mathbb{R}^{6\times 6}$. The update process is, 
\begin{align}
\left\{
\begin{aligned}
e_i &= y_i - \hat x_{i|i-1}, && \triangleright\text{estimation residual}
\\
S_i &= H P_{i|i-1} H^T + R, && \triangleright\text{residual covariance}\\
K_i &= P_{i|i-1} H^T S_i^{-1}, && \triangleright\text{Kalman gain}\\
\hat s_{i|i} &= \hat s_{i|i-1} + K_i e_i, && \triangleright\text{updated state}\\
P_{i|i} &= (I_{2d} - K_i H) P_{i|i-1}, && \triangleright\text{updated covariance}
\end{aligned}
\right.
\end{align}
where $ H=[I_3 \quad 0_3]$ is the measurement matrix, ${R}\in \mathbb{R}^{3 \times 3}$ is is a user-defined measurement noise covariance.
Finally, the additive velocity bias $b$ is extracted from the updated state $\hat s_{i|i}$ to produce the estimate of the adaptively corrected velocity defined in  \eqref{v_ekf}.

\subsection{Impedance Parameterization}
To tune the interaction dynamics along reproduced paths, the impedance characteristic must be parameterized. Inspired by \cite{kronander2015passive} and extending the formulation in \cite{yang2025nullspace}, we formulate a velocity tracking controller,
\begin{align}\label{imp_para}
    F_c &= D(\diamond) \, \dot p_c,
\end{align}
where $\dot p_c =  \dot y_{\text{msr}}-\dot y_{\text{ekf}}$, in that $\dot y_{\text{msr}}$ denotes the actual translational velocity measured from the robot.
$ D(\diamond) $ is the velocity
direction parameterized damping associated with the reproduced paths, it is designed to selectively tune the dissipation of energy in desired and undesired directions. In this work, the goal is to react to arbitrary external interactions with compliance while maintaining the tracking of demonstrated paths. Thus, let $\diamond := \dot y_{\text{ekf}}$, The parameterized damping is defined as,
\begin{align}\label{Damping}
    D(\diamond)&=U(\diamond)\Lambda U(\diamond)^T,
\end{align}
where matrix $U(\diamond) =[\hat e_1, \hat e_2, \hat e_3]  \in \mathbb{R}^{3\times 3}$ contains the estimated orthonormal principal axes $\hat e_1, \hat e_2, \hat e_3$, with $\hat e_1$ pointing in the desired direction of motion. Thus, let $\hat e_1 := \sfrac{\diamond}{||\diamond||}$, and the rest be represented as,
\begin{align}
\hat{ e}_{j} &:= 
      \frac{ v^{(j)}}{\max(\| v^{(j)}\|,\varepsilon)}, 
      \quad j=2,3,
\end{align}
where $v^{(j)} :=  e_k
      -\bigl(\hat{ e}_j^{\,T} e_k\bigr)\hat{ e}_j$,
      $\varepsilon \in \mathbb{R^+}$ denotes a numerical stability constant, $e_k$ denotes the $k^{th}$ canonical axis,  and for the $1^{st}$ canonical axis, $e_1 = [1, 0, 0]$. Additionally, in practice, the sign continuity of components in $U$ is ensured. $\Lambda = \text{diag}(\lambda_1,\lambda_2,\lambda_3)\in \mathbb{R}^{3\times 3}$ is a user-defined diagonal matrix with non-negative elements (typically 10–100). This is to adjust the perceived interaction behaviour. For example, setting $\lambda_1 = 0$ and others $>0$ allows for resisting the external force that results in deviations from the path tracking.

\subsection{A Self-learned Strategy for Surface Contact Task}
In the foregoing subsections, the proposed LfD scheme enabled the robot to reproduce a demonstrated path with both accuracy and compliance. For many manipulation tasks, however, the controller must also preserve a stable contact with the workpiece surface while following that path.  We exploit an intrinsic property of the demonstration phase: the human demonstrator is assumed to keep the EE tool in continuous contact with the target surface. Consequently, the local geometry of the surface is implicitly encoded in the DS, where no external contact or vision sensors are required. This observation allows the framework to handle arbitrarily oriented, tilted, or curved surfaces.

Taking advantage of the matrix $U$ derived in the previous subsection, under the contact-maintenance assumption, the third column $\hat e_3$
coincides with the outward surface normal. 
$\hat{ e}_3$ enabling a real-time estimate of the surface normal, we augment the velocity presented in \eqref{v_ekf} as,
\begin{align}\label{contact}
    \dot y_d =\dot y_{\text{mDS}} + g_c(\cdot) + b,
\end{align}
where $\dot y_d \in \mathbb{R}^3$ denotes the final generated velocity, with  $g_c(\cdot) = \alpha_c\hat{ e}_3$ creates an adjustable
velocity pointing to the surface  for contact maintaining. For example, a high $\alpha_c$ enforces tight contact, $\alpha_c = 0$ means the robot is tracking the demonstrated path with no contact force exerted.
Finally, we can write the total  control force by first reformulating the prior $\dot p_c$  in \eqref{imp_para} to $\dot p_c = \dot y_{\text{msr}}-\dot y_{{d}}$, then convert $F_c$ in \eqref{imp_para} to joint torque $\tau_c$ and substitute into the robot dynamics in \eqref{dynamicmodel}.

\begin{figure} [!ht]
   \begin{center}
\includegraphics[width=0.5\textwidth]
{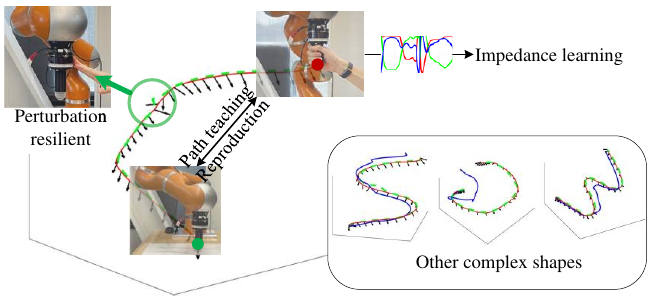}
   \end{center}
   \caption{Experimental procedure. The user teaches by physically guiding the EE along a 3D trapezoidal path. The green and red markers show the start and end points, respectively. External interaction is highlighted by a green circle, and the inset image shows the user’s applied “push” motion (arrow).}
   \label{Exp_setup}
\end{figure}

\begin{figure*} [!ht]
   \begin{center}
\includegraphics[width=0.83\textwidth]
{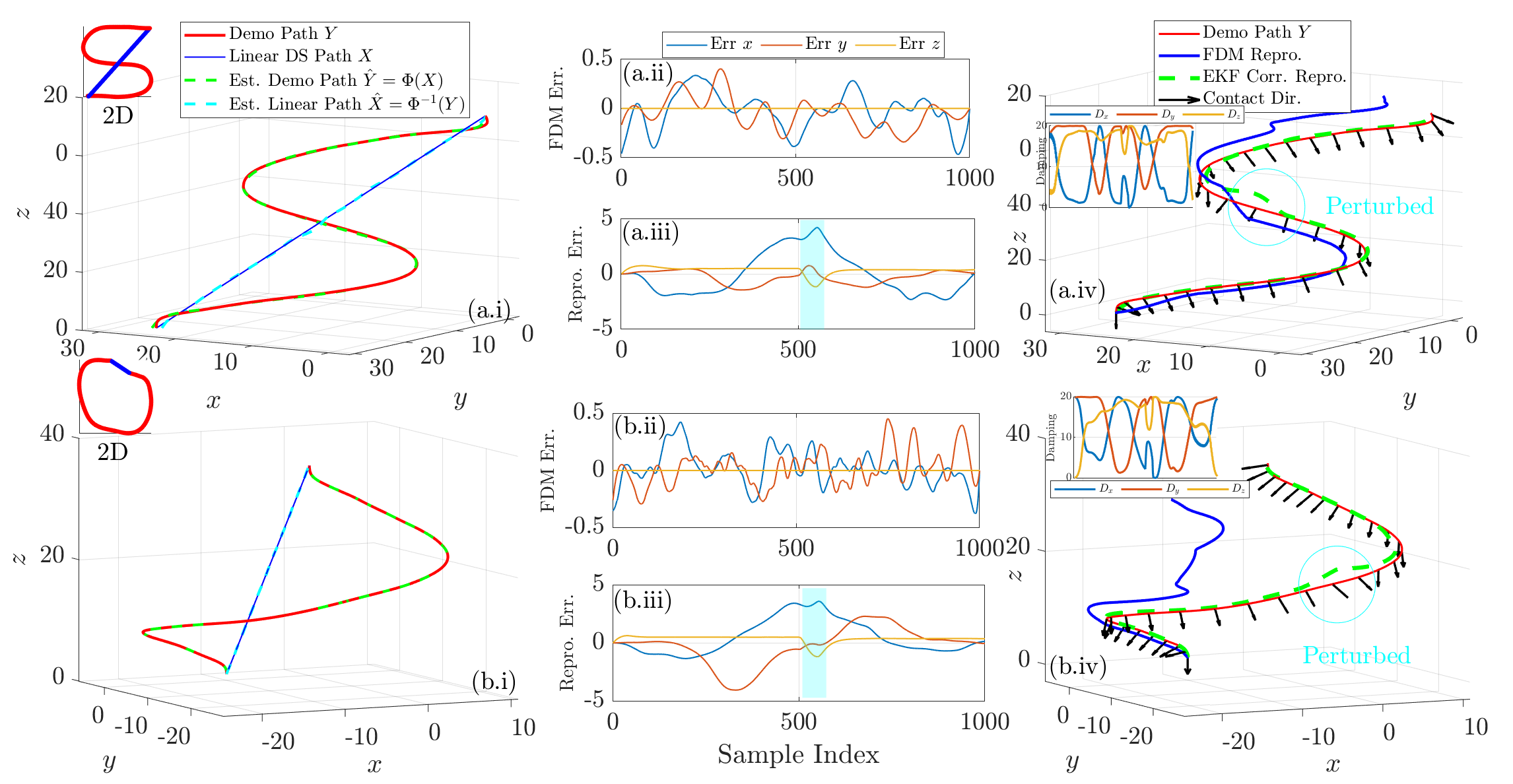}
   \end{center}
   \caption{Experiment result: (a,i) 3D FDM learned path, (a.ii–a.iv) estimation error of the learned path, reproduction error, and reproduction under perturbation for an `S'-shaped path; (b.i–b.iv) corresponding results for a near `O’-shaped path. Shaded areas indicate the period of interaction, with the corresponding impact highlighted in circles. All values are expressed in cm.}
\label{S+Blend}
\end{figure*} 

\begin{figure*} [!ht]
   \begin{center}
\includegraphics[width=0.83\textwidth]
{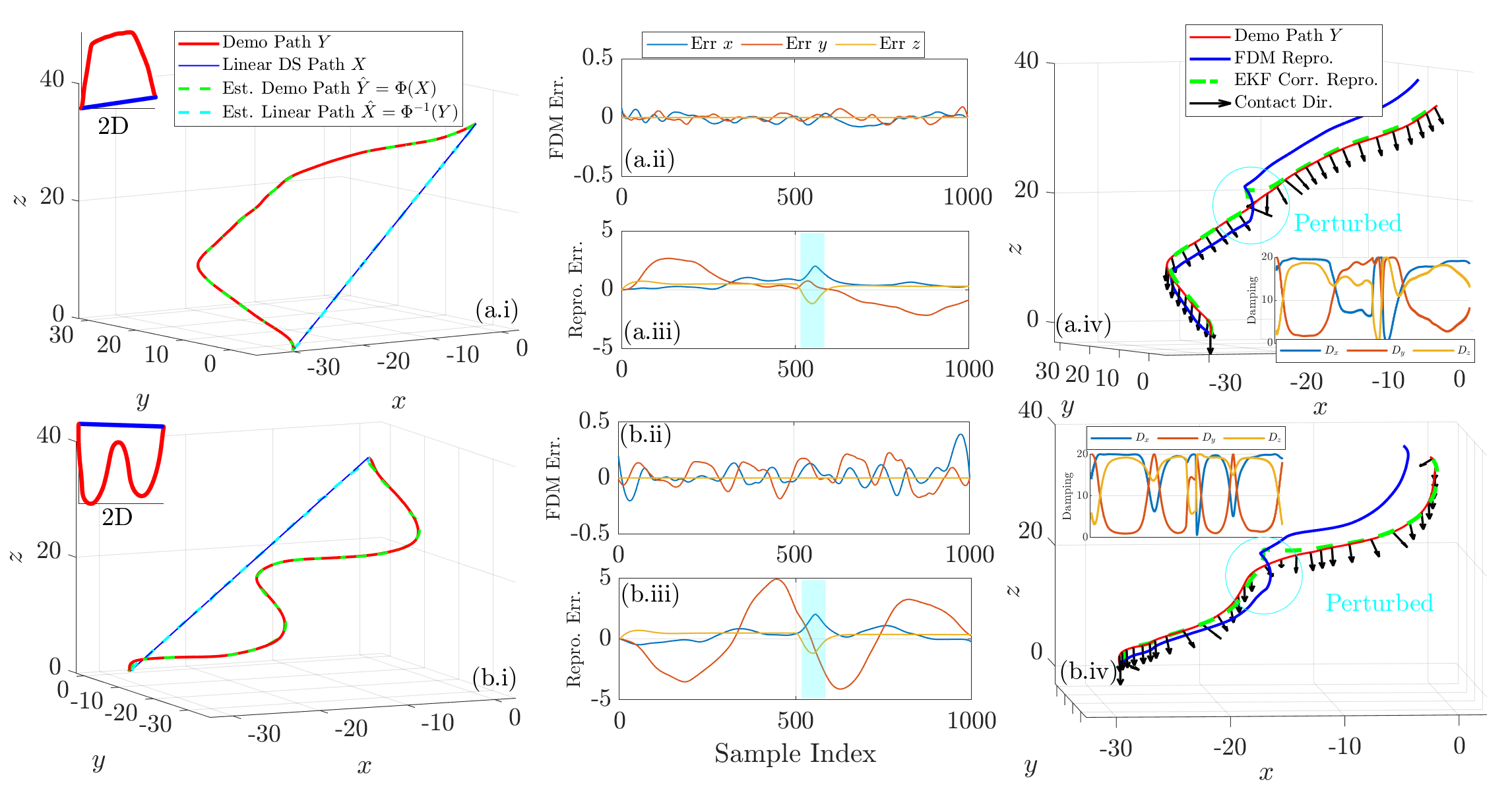}
   \end{center}
   \caption{Experiment result: (a,i) 3D FDM learned path, (a.ii–a.iv) estimation error of the learned path, reproduction error, and reproduction under perturbation for a trapezoid-shaped path; (b.i–b.iv) corresponding results for a `W’-shaped path. Shaded areas indicate the period of interaction, with the corresponding impact highlighted in circles. All values are expressed in cm.}
\label{W+trap}
\end{figure*} 

\section{Experimental EVALUATION}
\label{exp}
To evaluate the proposed method, we conduct experiments on a 7-DOF KUKA LWR IV+ robot manipulator as shown in Fig. \ref{Exp_setup}. 
The control algorithms are implemented on a remote Ubuntu PC with the Robot Operating System (ROS) framework. This remote PC establishes communication with the KUKA robot controller via UDP, utilizing the Fast Research Interface (FRI) with a sampling rate of 200 Hz.  Throughout the experiments,  the desired EE orientation is configured as $[\pi,0,\pi]^T$  (rad) to point towards the ground and controlled by a PD controller. The robot is controlled in joint impedance mode, in which the controlled torque is the only input to the robot, with
the control mode’s stiffness and damping interfaces disabled. Proper compensation for frictional ($\tau_f$) and gravitational torques is ensured.

In this experiment, we evaluate the learning and reproduction accuracy using 2D reference trajectories from the LASA Handwriting dataset \cite{KhansariZadeh2010LASA}. We extend from its original 2D formulation to 3D Cartesian space. Figs.~\ref{S+Blend} and \ref{W+trap} summarize the results. Subfigures (i) illustrate the FDM transformation by comparing the demonstrations ($Y$ in red) with their estimated  counterparts ($\hat Y$ in green), showing near-identical reconstruction. The inverse mapping $\hat{X}$ is  shown in light blue.  The associated estimation errors remain negligible (below 0.3 cm) and are reported in subfigures (ii). 

External perturbations are introduced to assess compliance and disturbance rejection. Results for the `S' and near `O'-shaped trajectories are shown in Fig.~\ref{S+Blend}.a–b, while trapezoidal and `W'-shaped paths are presented in Fig.~\ref{W+trap}.a–b.  The corrected reproduced trajectories (green dashed) closely follow the demonstrations (red), with reproduction errors generally under 2 cm, as shown in subfigures (iii) and (iv). Perturbation intervals are highlighted with a light blue shaded region in subfigure (iii), and recovery behaviour is illustrated in subfigure (iv) using light blue circles. In contrast, the DS-based velocity generation in~\eqref{DS} (dark blue) serves as a baseline method, exhibits visibly larger reproduction errors, which worsen with increasing path complexity and fails to recover under perturbations. Finally, subfigures (iv) also visualize the adaptive impedance parameters and the learned surface-contact strategy. Specifically, they depict the damping variations $D$ along the principal axes (shown in the inset plots), as well as the contact-maintaining behaviour indicated by the black arrows.

\section{Conclusion}
\label{Conclusion}
This paper presents an approach that integrates a 3D FDM algorithm with an EKF-DS-based motion generator to enable real-time single-shot LfD and precise motion reproduction. The proposed framework inherently supports perturbation resilience and surface-contact execution, making it particularly suitable for robot LfD applications that traditionally require extensive and repetitive human demonstrations.  By jointly ensuring compliance, perturbation recovery, and reproduction accuracy under varying interaction dynamics, the method provides a practical solution for interactive robotic tasks. Experimental validation on a KUKA LWR IV+ robot demonstrates the effectiveness and feasibility of the proposed approach, even when reproducing complex handwriting paths.

\newpage

\bibliographystyle{IEEEtran}
\bibliography{bib}
\end{document}